\documentclass[sigconf]{acmart}
\AtBeginDocument{%
  }
\usepackage{tabularx}
\usepackage{tikz}
\usepackage{pgf-pie}
\usepackage{subcaption}
\usepackage[table]{xcolor}
\usepackage{float}
\usepackage{placeins}
\usepackage{hyperref}
\usepackage{xurl}  % Better URL line-breaking

\setcopyright{cc}
\setcctype{by}
\copyrightyear{2026}
\acmYear{2026}
\acmDOI{10.1145/3836937.3836948}
\acmConference[ICAIL 2026]{21st International Conference on Artificial Intelligence and Law}{June 8--12,
  2026}{Singapore, Singapore}
\acmISBN{979-8-4007-2756-6/2026/06}

\begin{document}

%%
%% The "title" command has an optional parameter,
%% allowing the author to define a "short title" to be used in page headers.
\title{When Citations Mislead? A Claim-Level Benchmark for Legal Hallucination Detection}

%%
%% The "author" command and its associated commands are used to define
%% the authors and their affiliations.
%% Of note is the shared affiliation of the first two authors, and the
%% "authornote" and "authornotemark" commands
%% used to denote shared contribution to the research.
\author{M. Mikail Demir}
\affiliation{%
  \institution{University at Albany, SUNY}
  \city{Albany}
  \state{New York}
  \country{USA}
}
\email{mdemir@albany.edu}

\author{M. Abdullah Canbaz}
\affiliation{%
  \institution{University at Albany, SUNY}
  \city{Albany}
  \state{New York}
  \country{USA}
}
\email{mcanbaz@albany.edu}

%%
%% By default, the full list of authors will be used in the page
%% headers. Often, this list is too long, and will overlap
%% other information printed in the page headers. This command allows
%% the author to define a more concise list
%% of authors' names for this purpose.
\renewcommand{\shortauthors}{Demir and Canbaz}

%%
%% The abstract is a short summary of the work to be presented in the
%% article.
\begin{abstract}
Large language models are increasingly used in legal research and drafting, but they can still produce claims that sound convincing without being supported by the cited source. We introduce PARCEL, a benchmark for checking whether a legal claim is supported by the underlying authority. Using recent New York State Court of Appeals decisions, we build a dataset of 3,396 parenthetical-style claims labeled as Supported, Refuted, or Not Found. We cast this task as a three-way natural language inference problem and evaluate several state-of-the-art LLMs in a zero-shot setting. Although the strongest models reach up to 0.97 accuracy, the results also show an important weakness: models still incorrectly mark unsupported claims as supported, even when the full opinion text is provided. Across models, missing support is harder to detect than direct contradiction, and fabricated but plausible citations cause the largest drop in performance. Overall, PARCEL provides a practical benchmark for testing claim-level groundedness in legal RAG systems.
\end{abstract}

%%
%% The code below is generated by the tool at http://dl.acm.org/ccs.cfm.
%% Please copy and paste the code instead of the example below.
%%


%%
%% Keywords. The author(s) should pick words that accurately describe
%% the work being presented. Separate the keywords with commas.
%% A "teaser" image appears between the author and affiliation
%% information and the body of the document, and typically spans the
%% page.

%%
%% This command processes the author and affiliation and title
%% information and builds the first part of the formatted document.
\maketitle

\section{Introduction}
Large Language Models (LLMs) are rapidly reshaping how legal scholars and practitioners conduct research, draft documents, and interpret legal texts \cite{dahl2023largelegal,magesh2024hallucinationfree}. By enabling fast retrieval, summarization, and synthesis of large volumes of legal material, these models promise substantial efficiency gains. However, their deployment in high-stakes legal settings introduces serious risks. A well-documented failure mode of LLMs is hallucination which is the generation of fluent but factually incorrect or unsupported statements \cite{ji2023surveyNLG,huang2025surveyLLMHallucination}. In the legal domain, such errors have already resulted in professional and judicial consequences, including the citation of non-existent cases and unreliable legal authorities \cite{dahl2023largelegal,magesh2024hallucinationfree}.\footnote{For ongoing documentation of AI hallucination incidents in legal and professional settings, see practitioner-maintained trackers such as the AI Law Tracker of Polaris Lab (\url{https://www.polarislab.org/ai-law-tracker.html}) and Damien Charlotin’s compilation of hallucination cases (\url{https://www.damiencharlotin.com/hallucinations/)}.}

In practice, even a single unsupported statement, such as attributing a legal standard to the wrong case or overstating what a statute permits can materially alter legal analysis while remaining stylistically and rhetorically plausible \cite{reuters2025aihalluclaw}. Because legal writing is citation-driven, this failure often manifests as a confident claim paired with a citation that does not, in fact, support the asserted proposition \cite{magesh2024hallucinationfree}.

Research in Natural Language Processing (NLP) has long addressed related challenges through fact-checking, natural language inference, and textual entailment\cite{vlachosFactCheckingTask2014}. As LLMs are increasingly positioned as assistants or even automated agents within verification pipelines, the ability to reliably distinguish supported claims from unsupported or fabricated ones becomes central to safe use \cite{li2024halueval2,rawte2025factoid}. In this context, hallucination detection is not merely an auxiliary capability but a foundational requirement for aligning LLM behavior with evidence-sensitive tasks \cite{ji2023surveyNLG,huang2025surveyLLMHallucination}.

Despite growing interest in legal AI evaluation, existing benchmarks often fall short of capturing these risks. Many widely used datasets and tasks emphasize broad document understanding or general legal reasoning \cite{chalkidis2022lexglue,guha2023legalbench}. While valuable these studies do not reflect the granular, citation-driven verification processes that dominate real-world legal work \cite{magesh2024hallucinationfree,jiang2024wheredoeslegalaifail}. As a result, performance on these benchmarks may overstate a model’s readiness for deployment in practice-critical settings.

A particularly underexplored gap lies in the evaluation of hallucination detection within Retrieval-Augmented Generation (RAG) frameworks. Although RAG is commonly proposed as a mitigation strategy, empirical work shows that failures can arise from multiple components and still produce outputs that appear grounded \cite{jiang2024wheredoeslegalaifail}. While \citet{dahl2023largelegal} provide critical baseline evidence for open-domain hallucinations where models fail to recall or correctly identify real federal court cases, the core legal requirement for practitioners is often closed-domain claim-level verification against a specific, provided source.

To address this gap, we introduce \textbf{P}arenthetic\textbf{A}l \textbf{R}easoning and \textbf{C}laim \textbf{E}va\textbf{L}uation (\textsc{PARCEL}), a synthetic benchmark constructed from New York State Court of Appeals decisions issued between 2024 and 2025. From these opinions, we generate 3,000+ discrete, claim-level statements and label each instance under the standard Natural Language Inference (NLI) taxonomy: \textit{Supported} (entailment), \textit{Refuted} (contradiction), and \textit{Not Found} (neutral). 

A central feature of \textsc{PARCEL} is that it is grounded in real legal briefs submitted to the New York State Court of Appeals. We extract explanatory parenthetical citations from these briefs and use them as the basis for adversarial claim construction, allowing us to test whether models can detect subtle factual or logical deviations from the source text even when the resulting claims remain stylistically plausible and partially grounded.

Our contributions are twofold:
\begin{enumerate}
    \item We release a novel, adversarial benchmark for legal hallucination detection grounded in recent appellate case law, enabling claim-level evaluation under realistic citation-driven constraints.\footnote{The dataset is available at \url{https://github.com/mmikaildemir/PARCEL}.}
    \item We propose a systematic and reproducible pipeline for generating high-quality synthetic legal claims, including adversarial variants, to measure model groundedness and robustness to subtle logical and evidentiary pitfalls.
\end{enumerate}

We study the following research questions:
\begin{enumerate}
    \item How reliably can state-of-the-art LLMs detect fine-grained legal hallucinations in long-context RAG settings when claims are adversarially constructed yet stylistically realistic? 
    \item Which categories of adversarial legal hallucinations most strongly undermine model trustworthiness, and where do current models fail systematically?
    \item What factors explain systematic model failures, and can error analysis reveal limitations in current architectures for closed-book legal reasoning?
\end{enumerate}

In the sections that follow, we detail the dataset construction and claim-generation pipeline, and we present experiments demonstrating how adversarial, claim-level evaluation provides a reproducible framework for stress-testing the fact-verification assumptions underlying legal RAG systems.

\section{Related Work}

\subsection{Hallucinations in Legal Context}

Hallucinations in legal language models present distinct risks due to the normative, citation-driven nature of legal reasoning. Prior work has emphasized that legal hallucinations cannot be treated as generic factual errors, as they may appear stylistically correct while violating doctrinal, procedural, or evidentiary constraints. Dahl et al.~\cite{dahl2023largelegal} provide the first systematic study of this phenomenon, introducing a typology that distinguishes between failures internal to a model’s reasoning process and failures relative to external legal reality. Their findings demonstrate that hallucinations are prevalent across jurisdictions and model families, and that models often lack reliable self-awareness of these failures.

Within this framework, intrinsic hallucinations arise when a model is provided with authoritative legal materials but nonetheless misrepresents, distorts, or overgeneralizes facts contained within those documents. Such errors are especially concerning in legal workflows, where even subtle deviations, such as misstating the holding of a case or attributing a legal standard to an incorrect authority, can materially alter legal analysis. While broader NLP surveys categorize hallucinations along intrinsic–extrinsic or parametric–retrieval axes~\cite{ji2023surveyNLG,huang2025surveyLLMHallucination}, these distinctions take on heightened significance in law, where correctness is defined not only by surface factuality but by precise alignment with cited sources.

\subsection{Benchmarking LLMs in Law}

In parallel, substantial effort has been devoted to benchmarking LLM performance on legal tasks. Datasets such as LexGLUE~\cite{chalkidis2022lexglue} and LegalBench~\cite{guha2023legalbench} evaluate models across a range of legal reasoning and classification problems, including statute interpretation, case outcome prediction, and legal question answering. These benchmarks have been instrumental in advancing legal NLP, but they primarily assess document-level understanding or abstract legal knowledge.

Recent critiques argue that strong performance on such benchmarks may not translate to safe deployment in practice-critical settings. Recent literature has expanded the definition of legal hallucinations to include failures in procedural fidelity. For instance, \citet{dahlByebyeBluebookAutomating2025} \citeyear{dahlByebyeBluebookAutomating2025} found that even with in-context learning (ICL) of the rules, models reach only 77\% accuracy on Bluebook citation tasks, suggesting that legal form can be as difficult for LLMs to master as legal logic.In particular, these evaluations rarely test whether individual claims produced by a model are directly supported by a cited authority. Even RAG pipelines can fail in subtle ways, producing outputs that appear grounded while introducing unsupported inferences\cite{jiang2024wheredoeslegalaifail}. This gap highlights the need for evaluation frameworks that move beyond coarse-grained accuracy toward claim-level verification aligned with legal practice.

\begin{figure*}[!h]
\centering
\includegraphics[width=\textwidth]{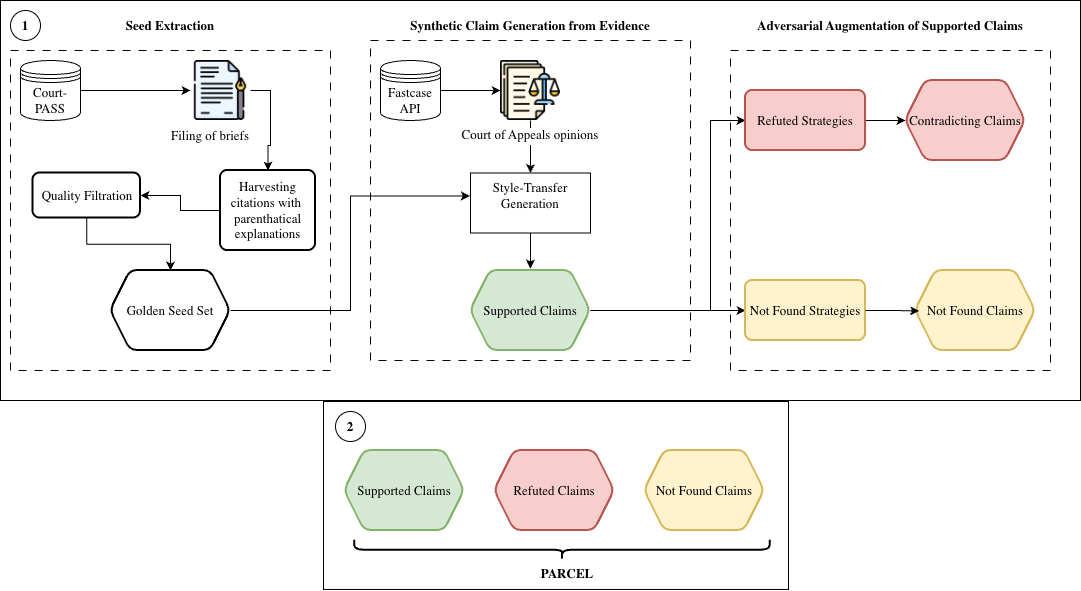}
\caption{Pipeline overview (seed extraction, synthetic claim generation and construction) of PARCEL.}
\label{fig:pipeline-image}
\end{figure*}

\subsection{Systematic Claim Generation for Fact-Checking Tasks}

Outside the legal domain, recent work has explored synthetic and adversarial claim generation as a tool for evaluating hallucination detection and factual consistency. Benchmarks such as HaluEval 2.0~\cite{li2024halueval2} and FACTOID~\cite{rawte2025factoid} generate claims that are entailed, contradicted, or unsupported by a reference text, enabling fine-grained analysis of model behavior under controlled conditions. These approaches demonstrate that carefully designed synthetic data can reveal systematic failure modes that are difficult to observe using naturally occurring benchmarks alone. Recent work in specialized domains further validates the utility of LLM-driven synthetic data for fact-checking. Decomposing source documents into atomic facts and systematically constructing claim-evidence entailment tables can significantly improve the performance of veracity verifiers in the medical domain\cite{zhangEnhancingHealthFactChecking2025a}.

However, existing claim-generation frameworks are largely domain-agnostic and do not account for the stylistic, logical, and citation-specific conventions of legal writing. Legal claims often embed implicit assumptions, doctrinal shortcuts, or jurisdictional nuances that make hallucinations particularly difficult to detect. Our work builds on this line of research by adapting systematic claim generation to the legal domain, grounding adversarial examples in real appellate decisions and aligning evaluation with the evidentiary standards of legal reasoning.

\section{Methodology}
Our methodology centers on explanatory parentheticals as the unit of analysis and combines harvesting from authentic appellate briefs with controlled synthetic generation to produce supported, refuted, and not-found claims aligned with long-context judicial decisions.

\subsection{Unit of Analysis: Explanatory Parentheticals}

Our benchmark is grounded in a specialized legal construct known as the explanatory parenthetical. According to the IndigoBook\footnote{\url{https://indigobook.github.io/versions/indigobook-2.0-rev2023-2.html}}, a public-domain implementation of the Bluebook system, explanatory parentheticals are optional additions used to “quickly convey important aspects” of judicial decisions~\cite{arredondolHARVESTINGUTILIZINGEXPLANATORY}. Structurally, they are appended to citations to clarify the relevance of the authority, often taking the form of a present participial phrase or a short statement.

We select explanatory parentheticals as the primary target for our fact-checking pipeline because they are ubiquitous across legal writing. Judges, clerks, and attorneys rely on parentheticals to convey substantive information about cited authorities in a compact form. Justice Ruth Bader Ginsburg has noted that a “first-rate brief” contains citations with “parenthetical explanations”~\cite{ginsburg1998remarks}. These explanations provide a direct window into the author’s professional interpretation of a source and highlight the precise relevance of a case to a specific legal proposition. Because parentheticals are authored by legally trained professionals to justify or illustrate a rule, they serve as a reliable proxy for the author’s intended legal logic.

From a technical perspective, explanatory parentheticals exhibit a highly regular linguistic structure governed by formal citation rules. They frequently begin with present participles such as \textit{holding}, \textit{finding}, or \textit{noting}, enabling high-precision automated harvesting at scale~\cite{arredondolHARVESTINGUTILIZINGEXPLANATORY}. At the same time, their usage is semantically diverse: parentheticals may summarize legal holdings, articulate reasoning rationales, describe procedural outcomes, or illustrate the concrete application of an abstract rule~\cite{voigtExplanatoryParentheticalsCan2013}. This balance of structural regularity and semantic richness makes them well-suited for fine-grained verification tasks.

\subsection{Task Definition}

We define \emph{Parenthetical Verification} as a ternary classification task. Given a judicial decision $D \in \mathcal{D}$ and a corresponding parenthetical claim $C \in \mathcal{C}$, the verification function $V$ maps the pair to a ternary label set $L = \{1, -1, 0\}$, representing entailment, contradiction, and neutral states:
\[
V: \mathcal{D} \times \mathcal{C} \rightarrow \{1, -1, 0\}.
\]

An \textbf{Entailment} ($1$) label indicates that the content of decision $D$ provides sufficient evidence to conclude that claim $C$ is a semantically accurate representation of the decision’s reasoning or factual content. A \textbf{Contradiction} ($-1$) label indicates that the decision directly refutes the claim. A \textbf{Neutral} ($0$) label indicates that the decision does not contain enough information to either support or refute the claim, corresponding to a \textit{Not Found} outcome in a legal fact-checking pipeline.

\subsection{Dataset Construction}

\subsubsection{Harvesting Parentheticals from Real Briefs}

To ensure stylistic realism, we first analyze explanatory parentheticals drawn from real appellate briefs. We retrieve 85 publicly available briefs filed before the New York State Court of Appeals via the CourtPASS system\footnote{\url{https://courtpass.nycourts.gov/}}. Using a targeted prompt, we instruct the google/gemini-3-flash-preview model to extract parenthetical explanations along with their cited cases, yielding over 700 candidate examples.

Manual inspection reveals two key patterns. First, many parentheticals are direct quotations, which we exclude because they merely restate source text rather than reflect the author’s interpretive reasoning. Second, a substantial portion of parentheticals are surface-level factual summaries that require minimal legal inference. To focus on non-trivial verification cases, we apply an LLM-as-a-judge approach to score each parenthetical by reasoning density. Parentheticals containing quotation marks or requiring only basic factual matching are filtered out, leaving 180 high-quality claims. These claims are used as few-shot exemplars in subsequent synthetic claim generation.

\subsubsection{Creating Synthetic Supported Claims from Judicial Decisions}

Because supported claims form the basis for generating distorted (refuted and not-found) claims, it is critical that they be realistic and stylistically aligned with professional legal writing. While harvested claims from real briefs are used in the first phase, relying on naturally occurring parentheticals alone would introduce severe class imbalance, as legal practitioners overwhelmingly cite authorities in a supportive manner. This imbalance can negatively affect a fact-checking benchmark’s ability to accurately classify claims against evidence~\cite{moeMitigatingClassImbalance2025}.

To address this issue, we synthetically generate supported claims directly from judicial decisions. An additional benefit of this approach is that it allows us to systematically instruct the model to produce claims reflecting diverse real-world usages of explanatory parentheticals. To this end, we design a usage-type rubric, summarized in Table~\ref{tab:usage-type-schema}, which captures common functional roles of parentheticals in legal writing.

\begin{table}[ht]
\caption{Usage type schema for synthetic claim generation.}
\vspace{2mm}  % Add space between caption and table

\label{tab:usage-type-schema}
\centering
\renewcommand{\arraystretch}{1.15}
\small
\vspace{-6mm}
\begin{tabularx}{\linewidth}{|>{\centering\arraybackslash}p{2.3cm}|>{\raggedright\arraybackslash}X|>{\raggedright\arraybackslash}X|}
\hline
\textbf{Usage Type} & \textbf{Definition} & \textbf{Example} \\
\hline
Holding rule & Summarizes the specific legal rule or final outcome established by the court. Focuses on the black letter law takeaway. & (holding that the crime of making a terroristic threat is a bail-qualifying offense) \\
\hline
Reasoning rationale & Explains the logical steps, policy justifications, or statutory interpretation used to reach the result. Focuses on why the court ruled that way. & (reasoning that the bail statute's disjunctive structure creates independent categories of qualifying offenses) \\
\hline
Facts procedure & Highlights key factual details or procedural history, often to support an analogy or distinction. Focuses on the narrative context. & (upholding bail where the defendant made repeated calls to a police emergency line threatening to shoot officers) \\
\hline
Treatment relational & Describes how the decision treats another specific authority or precedent. Focuses on the relationship between cases. & (distinguishing Matter of Perlbinder holdings because that case involved conflicting provisions with different levels of specificity) \\
\hline
Descriptive label & A compressed, non-participial description of the proceeding or context. Used to quickly flag the nature of the case without a full sentence. & (habeas corpus proceeding converting to declaratory judgment action) \\
\hline
\end{tabularx}
\end{table}

We retrieve Court of Appeals decisions issued between 2024 and 2025 and filter out decisions shorter than 2,000 characters to ensure sufficient contextual richness. Decisions triggering LLM safety filters, such as those containing detailed descriptions of sexual violence or homicide, are excluded. From the remaining corpus, we process 85 judicial decisions. Using few-shot prompting informed by harvested real-world examples, we instruct the model to generate up to ten supported claims per decision, ensuring that each usage type appears at least twice. This process yields a total of 849 supported claims.

\subsubsection{Creating Refuted and Not Found Claims}

From the supported claims, we generate refuted and not-found claims to simulate realistic hallucination scenarios. Not-found claims are particularly important given the documented tendencies of LLMs toward sycophancy and overgeneralization; they test model behavior when no supporting evidence exists to assess claim veracity.

Crucially, all adversarial claims are generated while conditioning on the full text of the judicial decision. We do not mutate claims in isolation. Instead, the LLM is provided with the complete opinion text and instructed to generate perturbations that remain topically consistent and vocabulary-aligned with the specific judge’s writing style.

Refuted claims are generated using the following strategies:
\begin{enumerate}
    \item \textbf{Distractor Entity Swap:} Swapping a named entity (e.g., a statute, co-defendant, or judge) with another entity appearing elsewhere in the same opinion.
    \item \textbf{Logic Reversal:} Inverting the legal outcome (e.g., \textit{affirmed} to \textit{reversed}) while preserving the court’s reasoning language.
    \item \textbf{Scope Contradiction:} Altering conditional qualifiers (e.g., changing “admissible if” to “admissible regardless of”), producing claims explicitly refuted by the opinion’s constraints.
    \item \textbf{Authority Swap:} Attributing the majority’s holding to a dissenting opinion, or vice versa.
\end{enumerate}

Not-found claims are generated using three strategies:
\begin{enumerate}
    \item \textbf{Phantom Citation:} Introducing a legally relevant but uncited precedent that does not appear in the decision.
    \item \textbf{Unaddressed Argument:} Generating a claim that the court declined to address a common defense or argument that was not actually discussed.
    \item \textbf{Missing Detail:} Inserting a plausible but omitted procedural fact (e.g., the specific sentence length, motion filing date, or trial judge's name) that is consistent with the case but absent from the decision text.
\end{enumerate}

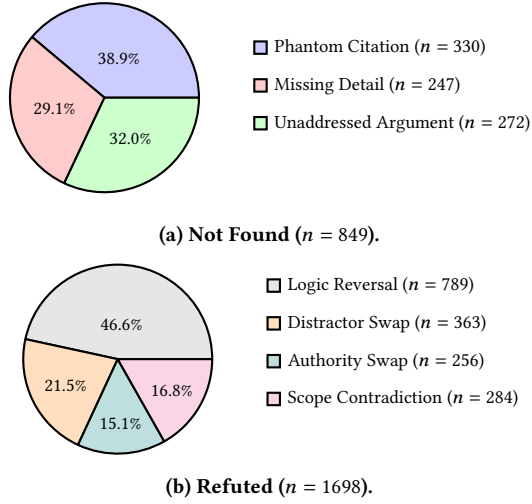
\begin{figure}[t]
\centering
\begin{subfigure}[b]{\columnwidth}
\centering
\footnotesize
\begin{tikzpicture}
\pie[
    radius=1.25,
    text=legend,
    after number=\%,
    color={blue!20, red!20, green!20}
]{
    38.9/Phantom Citation ($n=330$),
    29.1/Missing Detail ($n=247$),
    32.0/Unaddressed Argument ($n=272$)
}
\end{tikzpicture}
\caption{Not Found ($n=849$).}
\end{subfigure}

\vspace{0.5em}

\begin{subfigure}[b]{\columnwidth}
\centering
\footnotesize
\begin{tikzpicture}
\pie[
    radius=1.25,
    text=legend,
    after number=\%,
    color={gray!20, orange!25, teal!25, magenta!20}
]{
    46.6/Logic Reversal ($n=789$),
    21.5/Distractor Swap ($n=363$),
    15.1/Authority Swap ($n=256$),
    16.8/Scope Contradiction ($n=284$)
}
\end{tikzpicture}
\caption{Refuted ($n=1698$).}
\end{subfigure}
\caption{Distribution of adversarial generation strategies.}
\label{fig:adv-gen-strats}
\end{figure}

The resulting dataset contains 3,396 claims with a controlled 1:2:1 class balance across Supported, Refuted, and Not Found, respectively, which we report detailed breakdown in Figure~\ref{fig:adv-gen-strats}.

\subsection{Verification Setup}

To validate the dataset, we conduct a baseline reliability benchmark designed to assess model trustworthiness rather than raw accuracy. We evaluate whether state-of-the-art LLMs can distinguish between correctly grounded legal claims and subtly corrupted hallucinations in a zero-tolerance verification setting.

We formulate verification as a zero-shot Natural Language Inference task. Each model is provided with the full judicial decision as the premise and a single parenthetical claim as the hypothesis. A minimalist system prompt defines the task and label space, forcing the model to rely entirely on its internal representations, long-context processing, and implicit legal reasoning to perform claim verification.

The complete PARCEL construction process, from case selection through adversarial claim generation, is illustrated in Figure~\ref{fig:pipeline-image}. 

\begin{table*}[!h]
\centering
\small
\setlength{\tabcolsep}{3.5pt}
\renewcommand{\arraystretch}{1.15}
\caption{Model performance metrics (rounded to 2 decimals).}
\label{tab:model-metrics}
\resizebox{\textwidth}{!}{%
\begin{tabular}{lcc ccc ccc ccc}
\toprule
\textbf{Model} & \textbf{Overall Accuracy} & \textbf{Macro F1} & \multicolumn{3}{c}{\textbf{Supported}} & \multicolumn{3}{c}{\textbf{Refuted}} & \multicolumn{3}{c}{\textbf{Not Found}} \\
\cmidrule(lr){4-6} \cmidrule(lr){7-9} \cmidrule(lr){10-12}
 &  &  & \textbf{F1} & \textbf{Prec.} & \textbf{Rec.} & \textbf{F1} & \textbf{Prec.} & \textbf{Rec.} & \textbf{F1} & \textbf{Prec.} & \textbf{Rec.} \\
\midrule
gemini-3-flash-preview & 0.97 & 0.97 & 0.95 & 0.91 & 1.00 & 0.98 & 0.99 & 0.96 & 0.97 & 0.99 & 0.95 \\
kimi-k2-thinking & 0.95 & 0.94 & 0.94 & 0.88 & 0.99 & 0.97 & 0.97 & 0.97 & 0.91 & 0.98 & 0.86 \\
glm-4.7 & 0.94 & 0.93 & 0.94 & 0.89 & 0.99 & 0.96 & 0.94 & 0.99 & 0.88 & 0.99 & 0.79 \\
qwen3-235b-a22b-instruct-2507 & 0.93 & 0.92 & 0.91 & 0.84 & 0.99 & 0.96 & 0.97 & 0.96 & 0.89 & 0.97 & 0.82 \\
gemini-2.5-flash & 0.90 & 0.90 & 0.87 & 0.77 & 1.00 & 0.93 & 0.96 & 0.91 & 0.89 & 0.99 & 0.81 \\
gpt-oss-120b & 0.90 & 0.90 & 0.87 & 0.79 & 0.98 & 0.94 & 0.99 & 0.89 & 0.88 & 0.90 & 0.85 \\
llama-3.3-70b-instruct & 0.81 & 0.75 & 0.72 & 0.61 & 0.89 & 0.93 & 0.92 & 0.94 & 0.61 & 0.92 & 0.45 \\
\bottomrule
\end{tabular}%
}
\end{table*}

\section{Experimental Results}
\label{experimental-results}

In this section, we assess model trustworthiness for legal parenthetical verification under adversarially constructed claims. To this end, We structured our evaluation around three research questions. For RQ1 (overall reliability), we report baseline performance across seven state-of-the-art LLMs using standard classification metrics (Table~\ref{tab:model-metrics}), we quantify safety risk via False Entailment Rate (FER) (Table~\ref{tab:fer}), and we summarize error tendencies with an aggregated confusion matrix (Table~\ref{tab:confusion}). For RQ2 (strategy sensitivity), we break performance down by adversarial construction strategy for Refuted and Not Found claims (Tables~\ref{tab:refuted-strategy-by-model} and~\ref{tab:notfound-strategy-by-model}) to identify which perturbations most degrade reliability. For RQ3 (error diagnosis), we analyze “hard” claims, instances misclassified by multiple models, to surface systematic failure modes and evidence that parametric knowledge can override strict textual grounding (Table~\ref{tab:hard-claim-dist}).

\subsection{RQ1: Reliability of Hallucination Detection}

In this section, we evaluate the reliability of hallucination detection across seven state-of-the-art LLMs on \textsc{PARCEL}. In Table~\ref{tab:model-metrics}, we report overall accuracy, macro F1, and class-wise precision/recall/F1. gemini-3-flash-preview achieves the highest overall accuracy (97\%) and macro F1 (0.97), followed by kimi-k2-thinking (95\%) and glm-4.7 (94\%). As shown in Table~\ref{tab:model-metrics}, all models except llama-3.3-70b exceed 90\% accuracy, suggesting that modern LLMs can often perform claim-level legal verification when provided with the full long-context opinion.

However, Table~\ref{tab:model-metrics} also reveals that performance is not uniform across labels. In particular, the largest gap appears in the \textit{Not Found} class, which requires detecting the absence of supporting evidence. For example, llama-3.3-70b attains Not Found F1 = 0.61 with recall = 0.45, meaning it correctly identifies fewer than half of absent claims; by contrast, gemini-3-flash-preview reaches Not Found F1 = 0.97. This pattern, visible in the class-wise metrics in Table~\ref{tab:model-metrics}, indicates that absence detection, rather than contradiction recognition, is the primary bottleneck for legal hallucination detection.

At the same time, Table~\ref{tab:model-metrics} shows near-perfect recall on \textit{Supported} claims across models (0.89–1.00). High supported recall can be desirable, but it may also reflect an over-acceptance bias (i.e., a tendency to validate claims as supported even when evidence is missing). To quantify this safety risk directly, we use the False Entailment Rate (FER), reported in Table~\ref{tab:fer}. We define the ground-truth label for a pair $(D,C)$ as $y = V(D,C) \in \{1,-1,0\}$ and the model prediction as $\hat{y}=\hat{V}(D,C)$. Because predicting entailment asserts legal support, any case where $\hat{y}=1$ but $y \in \{-1,0\}$ constitutes a false entailment. We define:
\[
\text{FER} = P(\hat{y}=1 \mid y \in \{-1,0\}).
\]
To diagnose the source of these errors, we further decompose FER into two risk-specific components:
\begin{enumerate}
    \item \textbf{False Entailment on Refuted ($\text{FER}{-1}$):} The rate at which the model incorrectly claims support for a claim that is explicitly contradicted by the decision. $$ \text{FER}{-1} = P(\hat{y} = 1 \mid y = -1) $$
    \item \textbf{False Entailment on Neutral ($\text{FER}{0}$):} The rate at which the model incorrectly claims support for a claim that is absent from the decision. $$ \text{FER}{0} = P(\hat{y} = 1 \mid y = 0) $$
\end{enumerate}

In Table~\ref{tab:fer}, we observe a consistent asymmetry: $\text{FER}{0}$ is substantially larger than $\text{FER}{-1}$ across models. For instance, glm-4.7 achieves the lowest false entailment rate on refuted claims ($\text{FER}{-1}=1.18\%$), yet its false entailment rate on not-found claims rises to $\text{FER}{0}=9.55\%$. This pattern in Table~\ref{tab:fer} indicates that models are better at recognizing explicit contradictions than at recognizing when supporting evidence is absent, precisely the failure mode that produces plausible but unsupported legal assertions.

Table~\ref{tab:fer} also shows why accuracy alone can mask safety disparities. Although llama-3.3-70b reaches 0.81 overall accuracy (Table~\ref{tab:model-metrics}), it exhibits a severe absence-grounding failure with $\text{FER}{0}=48.76\%$ (Table~\ref{tab:fer}), meaning nearly half of Not Found claims are incorrectly validated as Supported. In contrast, Table~\ref{tab:fer} shows gemini-3-flash-preview achieves the lowest overall FER (3.47\%), making it the most consistently safe model under our adversarial verification setup. Notably, while kimi-k2-thinking slightly outperforms glm-4.7 in overall accuracy (Table~\ref{tab:model-metrics}), glm-4.7 attains a lower overall FER (Table~\ref{tab:fer}), reinforcing that FER captures trustworthiness properties that aggregate accuracy does not.

\begin{table}[!b]
\centering
\scriptsize
\setlength{\tabcolsep}{2.5pt}
\renewcommand{\arraystretch}{1.05}
\caption{False Entailment Rate (FER) by model.}
\label{tab:fer}
\resizebox{\columnwidth}{!}{%
\begin{tabular}{lccc}
\toprule
\textbf{Model} & \textbf{FER (Overall)} & \textbf{($\text{FER}{-1}$) (Refuted)} & \textbf{($\text{FER}{0}$) (Not Found)} \\
\midrule
gemini-3-flash-preview & 3.47\% & 3.37\% & 3.67\% \\
glm-4.7 & 3.97\% & 1.18\% & 9.55\% \\
kimi-k2-thinking & 4.36\% & 2.41\% & 8.25\% \\
qwen3-235b & 6.17\% & 2.95\% & 12.60\% \\
gpt-oss-120b & 8.96\% & 6.73\% & 13.43\% \\
gemini-2.5-flash & 9.98\% & 9.14\% & 11.66\% \\
llama-3.3-70b & 19.21\% & 4.42\% & 48.76\% \\
\bottomrule
\end{tabular}%
}
\end{table}

\begin{table}[b]
\centering
\small
\setlength{\tabcolsep}{6pt}
\renewcommand{\arraystretch}{1.2}
\caption{Aggregate confusion matrix (all models), reported as percentage within each ground-truth class.}
\label{tab:confusion}
\begin{tabular}{lccc}
\toprule
 & \multicolumn{3}{c}{\textbf{Predicted}} \\
\cmidrule(lr){2-4}
\textbf{Ground Truth} & \textbf{Supported} & \textbf{Refuted} & \textbf{Not Found} \\
\midrule
Supported & \cellcolor{blue!98}\textcolor{white}{97.8\%} & \cellcolor{blue!2}2.0\% & \cellcolor{blue!0}0.2\% \\
Refuted & \cellcolor{blue!5}4.3\% & \cellcolor{blue!94}\textcolor{white}{94.3\%} & \cellcolor{blue!1}1.4\% \\
Not Found & \cellcolor{blue!15}15.4\% & \cellcolor{blue!5}5.4\% & \cellcolor{blue!79}\textcolor{white}{79.2\%} \\
\bottomrule
\end{tabular}
\end{table}

Finally, in Table~\ref{tab:confusion}, we show an aggregated confusion matrix (percentages within each ground-truth class) that summarizes error directionality across all models. Table~\ref{tab:confusion} confirms the dominant failure mode: 15.4\% of Not Found claims are misclassified as Supported, whereas only 4.3\% of Refuted claims are misclassified as Supported. Together, the class-wise metrics in Table~\ref{tab:model-metrics}, the safety analysis in Table~\ref{tab:fer}, and the error directionality in Table~\ref{tab:confusion} consistently indicate that detecting evidentiary absence, not contradiction, is the main remaining challenge for reliable hallucination detection in legal verification.

\subsection{RQ2 (Failure Modes by Adversarial Strategy)}

In this section, we analyze how different adversarial claim-generation strategies affect model reliability. We focus on strategy-specific accuracy to identify which types of hallucinations are most difficult to detect and therefore pose the greatest risk in legal verification settings.

We first examine refuted claims. In Table~\ref{tab:refuted-strategy-by-model}, we report accuracy for four contradiction strategies. Overall, models perform strongly on contradiction detection, with average accuracy ranging from 0.86 for Authority Swap to 0.97 for Scope Contradiction. glm-4.7 exhibits near-perfect performance across all refuted strategies, achieving 1.00 accuracy on Scope Contradiction. These results suggest that explicit logical inconsistencies, such as reversed holdings or violated conditions, are comparatively easy for models to detect when the contradiction is directly encoded in the text.

However, performance drops for strategies that require discourse-level attribution rather than local logical comparison. Authority Swap is consistently the most challenging refuted strategy: gemini-2.5-flash and gpt-oss-120b both drop to 0.77 accuracy, indicating difficulty distinguishing majority opinions from dissents or concurrences. This pattern highlights a structural limitation in modeling legal discourse roles, even when the relevant information is present in the premise.

\begin{table}[t]
\centering
\tiny
\setlength{\tabcolsep}{1.0pt}
\renewcommand{\arraystretch}{1.0}
\caption{Accuracy metric for refuted-claim strategies (rounded to 2 decimals).}
\label{tab:refuted-strategy-by-model}
\resizebox{0.9\columnwidth}{!}{%
\begin{tabular}{lcccc}
\toprule
\textbf{Model} & \textbf{Distr. Swap} & \textbf{Logic Rev.} & \textbf{Scope Contr.} & \textbf{Auth. Swap} \\
\midrule
glm-4.7 & 0.97 & 0.99 & 1.00 & 0.98 \\
gemini-2.5-flash & 0.94 & 0.91 & 0.95 & 0.77 \\
gemini-3-flash-preview Preview & 0.98 & 0.97 & 0.98 & 0.86 \\
gpt-oss-120b & 0.80 & 0.95 & 0.93 & 0.77 \\
kimi-k2-thinking & 0.96 & 0.98 & 0.99 & 0.88 \\
llama-3.3-70b-instruct & 0.81 & 0.98 & 0.97 & 0.93 \\
qwen3-235b-a22b-instruct & 0.94 & 0.97 & 0.98 & 0.89 \\
\midrule
\textbf{Average} & \textbf{0.91} & \textbf{0.96} & \textbf{0.97} & \textbf{0.86} \\
\bottomrule
\end{tabular}%
}

\end{table}

% \FloatBarrier
\begin{table}[t]
\centering
\tiny
\setlength{\tabcolsep}{1.3pt}
\renewcommand{\arraystretch}{1.0}
\caption{Accuracy metric for not found-claim strategies (rounded to 2 decimals).}
\label{tab:notfound-strategy-by-model}
\resizebox{0.9\columnwidth}{!}{%
\begin{tabular}{lccc}
\toprule
\textbf{Model} & \textbf{Phantom Cit.} & \textbf{Missing Det.} & \textbf{Unaddr. Arg.} \\
\midrule
glm-4.7 & 0.73 & 0.83 & 0.84 \\
gemini-2.5-flash & 0.80 & 0.97 & 0.66 \\
gemini-3-flash-preview Preview & 0.95 & 0.96 & 0.93 \\
gpt-oss-120b & 0.80 & 0.90 & 0.88 \\
kimi-k2-thinking & 0.86 & 0.91 & 0.82 \\
llama-3.3-70b-instruct & 0.24 & 0.45 & 0.72 \\
qwen3-235b-a22b-instruct & 0.81 & 0.77 & 0.90 \\
\midrule
\textbf{Average} & \textbf{0.72} & \textbf{0.81} & \textbf{0.80} \\
\bottomrule
\end{tabular}%
}
\end{table}

We next analyze not-found claims, which test a model’s ability to detect the absence of supporting evidence. In Table~\ref{tab:notfound-strategy-by-model}, we report accuracy for Phantom Citation, Missing Detail, and Unaddressed Argument strategies. Across all models, absence detection is substantially harder than contradiction detection. Average accuracy ranges from 0.72 for Phantom Citation to 0.81 for Missing Detail, confirming that fabricated but legally plausible content poses a greater challenge than explicitly false statements.

Phantom Citation is the most difficult adversarial strategy. For example, llama-3.3-70b achieves only 0.24 accuracy on this category, frequently validating claims that introduce legally relevant but uncited authorities. In contrast, gemini-3-flash-preview remains robust across all not-found strategies, with accuracy between 0.93 and 0.96. These results indicate that strong performance on contradiction detection does not necessarily translate to reliable absence detection, a distinction that is critical in legal RAG settings.

Taken together, Tables~\ref{tab:refuted-strategy-by-model} and~\ref{tab:notfound-strategy-by-model} reveal a clear difficulty hierarchy across adversarial strategies. Models perform best on Scope Contradiction and Logic Reversal, followed by Distractor Swap and Authority Swap, and perform worst on absence-based strategies, especially Phantom Citation. This hierarchy suggests that hallucinations involving missing or fabricated authority represent the most persistent and consequential failure mode for current LLM-based legal verification systems.

\begin{table}[t]
\centering
\small
\caption{Hard Claim Distribution. Number of claims where $\ge k$ models failed.}
\label{tab:hard-claim-dist}
\begin{tabular}{lcc}
\toprule
\textbf{$k$ (min failures)} & \textbf{Claims} & \textbf{\% of 3396} \\
\midrule
$k \ge 3$ & 290 & 8.54\% \\
$k \ge 4$ & 158 & 4.65\% \\
$k \ge 5$ & 75 & 2.21\% \\
$k \ge 6$ & 27 & 0.80\% \\
$k = 7$ (all failed) & 8 & 0.24\% \\
\bottomrule
\end{tabular}
\end{table}

\subsection{RQ3 (Systematic Failures and Error Diagnosis)}

To identify systematic failure modes beyond individual model weaknesses, we analyze claims that multiple models misclassify simultaneously. In Table~\ref{tab:hard-claim-dist}, we report the distribution of “hard” claims, defined as instances misclassified by at least $k$ out of seven models. We observe that 290 claims (8.54\% of the dataset) are misclassified by at least three models, while only 8 claims (0.24\%) defeat all seven models. This sharp decline indicates that most errors are not universal but model-specific.

This distribution has practical implications. Because the majority of hard claims are correctly classified by at least some models, ensemble-based verification strategies, such as majority voting or verifier cascades, could substantially reduce hallucination risk. Only a very small fraction of claims remain universally difficult, suggesting that current models collectively possess the necessary capabilities for legal claim verification, even if individual systems exhibit blind spots.

To understand why certain claims remain challenging, we manually examined the 27 instances where at least six models fail. While some errors arise from extremely subtle perturbations, such as minimal entity swaps or lightly altered procedural details, we observe a recurring and more concerning pattern: models often rely on parametric legal knowledge rather than the provided textual evidence. In particular, models sometimes validate claims that are legally accurate in the abstract but unsupported by the specific judicial decision.

For example, a not-found claim describing an appeal as involving “leaving the scene of a personal injury accident” is frequently labeled as Supported, even though the opinion text only references a violation of Vehicle and Traffic Law \S 600 (2) (a) without using this descriptive language. Consider the following not found claim:

\begin{center}
\fbox{%
\begin{minipage}{0.95\linewidth}
\ttfamily\footnotesize
\noindent (appeal from a criminal conviction involving leaving the scene of a personal injury accident)
\end{minipage}%
}
\end{center}

The critical detail "leaving the scene of a personal injury accident" does not appear anywhere in the judicial decision. However the decision clearly indicates that there is a Vehicle and Traffic Law § 600 (2) (a) violation:

\begin{center}
\fbox{%
\begin{minipage}{0.95\linewidth}
\ttfamily\footnotesize
\noindent (Order affirmed. Viewing the evidence in the light most favorable to the People, "there is a valid line of reasoning and permissible inferences from which a rational jury could have found" that the defendant violated Vehicle and Traffic Law § 600 (2) (a) (People v Rouse, 34 N.Y.3d 269, 274 [2019]).)
\end{minipage}%
}
\end{center}
In this example, the adversarial claim restates a statutory code in plain English. Although legally accurate, terms like "leaving the scene" and "personal injury" do not appear in the original text. Models that label the claim as Supported show knowledge leakage, letting pre-trained legal knowledge override the NLI task’s strict textual grounding. Even if the plain-English claim reflects the statute’s real-world meaning, it is not in the premise, so marking it as supported violates the verification task’s grounding requirement.

This behavior reflects an inverse contamination effect. Rather than benefiting from memorized benchmark content, models’ pre-trained legal knowledge interferes with closed-book verification by overriding textual constraints. This finding aligns with prior work on parametric–contextual knowledge conflicts in LLMs \cite{longpre2021entity,xie2024adaptive,neeman2023disentqa} and extends concerns about legal benchmark contamination \cite{fei2024lawbench}. In high-stakes legal settings, this tendency is particularly risky: a claim may be legally correct in general, yet still constitute a hallucination if it is not supported by the cited authority. Our analysis shows that this distinction remains a fundamental challenge for current legal language models.

% This observation represents an inverse contamination effect. Rather than inflating scores by memorizing benchmark answers, models' parametric knowledge of real legal cases causes them to incorrectly validate fabricated details that happen to be factually true outside the premise. This aligns with prior research on knowledge conflicts in LLMs \cite{longpre2021entity,xie2024adaptive,neeman2023disentqa}. Models can over-rely on parametric knowledge when it conflicts with provided context \cite{longpre2021entity}, and may default to internal knowledge even when presented with contradictory evidence \cite{xie2024adaptive}. Current architectures also struggle to disentangle parametric and contextual knowledge in counterfactual settings \cite{neeman2023disentqa}. Similar contamination concerns have been reported in the legal domain \cite{fei2024lawbench}. Our finding extends this literature by demonstrating that for adversarial closed-book NLI, external knowledge can paradoxically hurt performance.

\section{Discussion}

Our results demonstrate that state-of-the-art LLMs can perform claim-level legal verification with high apparent accuracy, even under a strict zero-shot setup and with long-form judicial opinions as input. This finding suggests that modern long-context architectures are well-suited to “needle-in-the-haystack” reasoning tasks that require aligning fine-grained claims with dense legal texts. Several of the strongest-performing models also incorporate explicit or implicit reasoning mechanisms, which likely support more structured comparison between a hypothesis and an extended premise. Taken together, these results indicate meaningful progress toward practical AI-assisted legal fact-checking.

However, our analysis also shows that aggregate accuracy substantially overstates model trustworthiness in legal settings. When examined through a safety-oriented lens, models frequently over-validate unsupported claims, particularly when those claims are legally plausible or align with widely known doctrines. This behavior reflects a core risk in real-world legal practice: a hallucinated parenthetical that “sounds right” and cites a real authority may evade detection during routine review, despite lacking textual support. By isolating claim-level verification and introducing adversarial but stylistically realistic perturbations, PARCEL exposes a gap between surface competence and verifiable grounding that is not captured by existing legal NLP benchmarks.

A key finding across models is that absence detection remains the dominant bottleneck. Models are consistently better at identifying explicit contradictions than at recognizing when relevant information is missing from the source text. This asymmetry is especially problematic for Retrieval-Augmented Generation pipelines, where hallucinations often take the form of fabricated details or uncited authorities rather than direct falsehoods. Our False Entailment Rate analysis shows that these errors represent the most consequential failure mode, as they correspond to confidently asserted legal support where none exists.

We further observe that long-context access alone does not resolve this issue. Even when provided with the full judicial opinion, models frequently rely on parametric legal knowledge, validating claims that are legally accurate in the abstract but unsupported by the specific decision at hand. This tension between general legal understanding and strict textual grounding highlights a fundamental limitation of current architectures in closed-book verification settings. In law, where authority derives from what a court actually held, not from what is generally true, this distinction is critical.

Finally, our analysis of hard claims reveals both a challenge and an opportunity. While a small subset of claims systematically defeats multiple models, the rapid decline in universally failed cases suggests that model diversity can be leveraged as a practical mitigation strategy. Ensemble-based verification or secondary checking mechanisms may therefore offer a viable path toward reducing hallucination risk in applied legal systems, even before architectural advances fully resolve grounding limitations.

\section{Conclusion and Future Work}

PARCEL is motivated by a practical legal review task: determining whether a specific proposition is actually supported by the cited authority. By grounding evaluation in explanatory parentheticals and recent appellate decisions, we construct a benchmark that reflects how legal claims are made, justified, and reviewed in practice. Our results show that although contemporary LLMs often perform strongly on this task, they remain vulnerable to precisely the hallucinations that matter most in legal settings, including fabricated citations, missing details, and plausible but unsupported abstractions.

These findings underscore the need for evaluation frameworks that prioritize groundedness and safety over aggregate accuracy. PARCEL provides a reproducible foundation for studying those failures under realistic, practice-driven conditions, and it opens several directions for future work, including uncertainty-aware evaluation, stronger citation verification, and verifier- or ensemble-based mitigation strategies \cite{metropolitanskyEffectiveExtractionEvaluation2025}. We hope this benchmark supports the development of AI systems that legal professionals can trust not only to sound correct, but to be demonstrably grounded in the law.

\section{Acknowledgments}

Research reported in this publication was supported in part by SUNY System Administration using the SUNY AI Platform.

\bibliographystyle{ACM-Reference-Format}
\bibliography{acmart}

%%
%% If your work has an appendix, this is the place to put it.

\appendix
\section{Prompts and Dataset}
All prompts used in this study, along with the dataset, are available at \url{https://github.com/mmikaildemir/PARCEL}. The repository includes templates for claim generation (Supported, Refuted, Not Found) and model evaluation.

\end{document}